\documentclass[11pt,a4paper]{article}
\usepackage[hyperref]{acl2021}
\usepackage{times}
\usepackage{latexsym}

\usepackage{microtype}

\usepackage{algorithm}
\usepackage{algorithmicx}
\usepackage{algpseudocode}
\usepackage{amsmath}
\usepackage{graphics}
\usepackage{epsfig}
\usepackage{setspace} 
\usepackage{booktabs}  
\usepackage{threeparttable}  
\usepackage{multirow}

\usepackage{graphicx}%图片头文件
\usepackage{float} %指定图片位置
\usepackage{subfigure}%并排子图 共享标题 有子标题

\usepackage{amsmath}
\usepackage{color, xcolor}
\usepackage{amssymb}% http://ctan.org/pkg/amssymb
\usepackage{pifont}% http://ctan.org/pkg/pifont
\newcommand{\cmark}{\ding{51}}%
\newcommand{\xmark}{\ding{55}}%

\aclfinalcopy % Uncomment this line for the final submission
\def\aclpaperid{1258} %  Enter the acl Paper ID here

\title{Meta-Learning Adversarial Domain Adaptation Network for Few-Shot Text Classification}

\author{ChengCheng Han\textsuperscript{1}\ \ \ \ Zeqiu Fan\textsuperscript{1}\ \ \ \ Dongxiang Zhang\textsuperscript{2}\ \ \ \ Minghui Qiu\textsuperscript{3} \\
\bf{Ming Gao\textsuperscript{1\thanks{\ \ Corresponding author}}}\ \ \ \ \bf{Aoying Zhou\textsuperscript{1}} \\
    \textsuperscript{1}School of Data Science and Engineering, East China Normal University \\  \textsuperscript{2}College of Computer Science and Technology, Zhejiang University \\ \textsuperscript{3}Alibaba Group \\
    \texttt{\{51195100009,51195100007\}@stu.ecnu.edu.cn}  \\
              \texttt{zhangdongxiang@zju.edu.cn} \\ \texttt{minghui.qmh@alibaba-inc.com} \\ \texttt{mgao@dase.ecnu.edu.cn} \\
              \texttt{ayzhou@sei.ecnu.edu.cn} }

\date{}

\begin{document}
\maketitle
\begin{abstract}

Meta-learning has emerged as a trending technique to tackle few-shot text classification and achieved state-of-the-art performance. However, existing solutions heavily rely on the exploitation of lexical features and their distributional signatures on training data, while neglecting to strengthen the model's ability to adapt to new tasks. In this paper, we propose a novel meta-learning framework integrated with an adversarial domain adaptation network, aiming to improve the adaptive ability of the model and generate high-quality text embedding for new classes. Extensive experiments are conducted on four benchmark datasets and our method demonstrates clear superiority over the state-of-the-art models in all the datasets. In particular, the accuracy of 1-shot and 5-shot classification on the dataset of 20 Newsgroups is boosted from $52.1\%$ to $59.6\%$, and from $68.3\%$ to $77.8\%$, respectively\footnote{The source code of the paper is available at \url{https://github.com/hccngu/MLADA}.}.

\end{abstract}

\section{Introduction}

Few-shot text classification \cite{DBLP:conf/naacl/YuGYCPCTWZ18,DBLP:conf/emnlp/GengLLZJS19} is a task in which a model will be adapted to predict new classes not seen in training. For each of these new classes, we only have a few labeled examples. To be specific, we are given lots of training data with a set of classes $\mathcal{Y}_{train}$. After training, our goal is to get accurate classification results on the testing data with a set of new classes $\mathcal{Y}_{test}$, which is disjoint to $\mathcal{Y}_{train}$. Only a small labeled support set will be available in the testing stage. If the support set contains \textit{K} labeled examples for each of the \textit{N} unique classes, we refer to the task as a \textit{N}-way \textit{K}-shot classification. 

Existing approaches for few-shot text classification mainly fall into two categories: (1) transfer-learning based methods \cite{DBLP:conf/acl/RuderH18, DBLP:journals/access/PanHGY19, DBLP:conf/coling/GuptaTO20}, which aim to transfer knowledge learned from a task to a new task or leverage general-domain pretraining and fine-tuning techniques for few-shot classification.
% However, transfer learning is better at solving the problem caused by the different distribution of the source domain and the target domain, which is less effective on few-shot text classification. 
%An more effective solution for few-shot text classification is 
(2) meta-learning based methods~\cite{DBLP:journals/corr/abs-1805-07722,DBLP:conf/naacl/YuGYCPCTWZ18,DBLP:conf/emnlp/GengLLZJS19,DBLP:conf/acl/GengLLSZ20,DBLP:conf/iclr/BaoWCB20}, which aim to learn generic information (meta-knowledge) by recreating training episodes, so that it can classify new classes through only a few labeled examples. Among these methods, \citet{DBLP:conf/iclr/BaoWCB20} leveraged distributional signatures (e.g. word frequency and information entropy) to train a model within a meta-learning framework, and achieved state-of-the-art performance. However, the method pays more attention to statistical information and ignores other implicit information such as correlation between words. Furthermore, existing meta-learning methods heavily rely on the exploitation of lexical features and their distributional signatures on training data, while neglecting to strengthen the model's ability to adapt to new tasks.

In this paper, we propose an adversarial domain adaptation network to enhance meta-learning framework, with the objective of improving the model's adaptive ability for new tasks in new domains. We first utilize two neural networks competing against each other, separately playing the roles of a domain discriminator and a meta-knowledge generator. The adversarial network is able to strengthen the adaptability of the meta-learning architecture. Moreover, we aggregate transferable features generated by the meta-knowledge generator with sentence-specific features to produce high-quality sentence embeddings. Finally, we utilize a ridge regression classifier to obtain final classification results. To the best of our knowledge, we are the first to combine adversarial domain adaptation with meta-learning for few-shot text classification.

We evaluate our model on four popular datasets for few-shot text classification. Experimental results demonstrate that our method outperforms state-of-the-art models in all datasets, for both in 1-shot and 5-shot classification tasks. Especially on the 20 Newsgroups dataset, our model outperforms DS-FSL \cite{DBLP:conf/iclr/BaoWCB20} by $7.5\%$ in 1-shot classification and $9.5\%$ in 5-shot classification. In addition, we conduct visualization analysis to verify the adaptability of our model and capability to recognize important lexical features for unseen classes.

\section{Related Work}

The mainstream approaches for few-shot text classification are based on  meta-learning or transfer learning. In this section, we first briefly introduce the preliminary background of these two technologies, and then review how they are applied to support few-shot text classification. %Finally, we address the relationship between our approach and these two techniques.

% \paragraph{Transfer learning}~Transfer learning aims to leverage knowledge from a related domain (a.k.a. source domain) to improve the learning performance and reduce the reliance on the number of labeled examples required in a target domain. The transfer learning approaches can be categorized into four groups: instance-based, feature-based, parameter-based, and relational-based approaches. Instance-based methods \cite{DBLP:conf/nips/HuangSGBS06, DBLP:conf/nips/SugiyamaNKBK07} refer to re-weight some labeled data in the source domain for use in the target domain. Feature-based methods \cite{DBLP:journals/pami/XiaoG15, DBLP:journals/iet-cvi/ShiZLZ19} transform the original features to create a new feature representation. Domain adaptation \cite{DBLP:journals/jmlr/GaninUAGLLML16, DBLP:conf/cvpr/TzengHSD17} is a feature-based method, which aims to bridge the gap between the source and target domains by learning domain-invariant feature representations. Parameter-based methods \cite{DBLP:journals/tcyb/WangLWLZZZ20} is to directly share the parameters of the source learner to the target learner, including pre-trained language models \cite{DBLP:conf/naacl/DevlinCLT19,DBLP:conf/nips/YangDYCSL19}. Relational-based approaches \cite{DBLP:journals/remotesensing/QinYLSL19} mainly focus on the problems in relational domains. Such approaches transfer the logical relationship or rules learned in the source domain to the target domain.

\paragraph{Meta-learning}~Meta-learning, also known as ``learning to learn'', refers to improving the learning ability of a model through multiple training episodes so that it can learn new tasks or adapt to new environments quickly with a few training examples. Existing approaches mainly fall into two categories: (1) Optimization-based methods , including developing a meta-learner as optimizer to output search steps for each learner directly \cite{DBLP:conf/nips/AndrychowiczDCH16,DBLP:conf/iclr/RaviL17,DBLP:conf/iclr/MishraR0A18,DBLP:conf/iclr/GordonBBNT19} and learning an optimized initialization of model parameters, which can be later adapted to new tasks by a few steps of gradient descent \cite{DBLP:conf/icml/FinnAL17,DBLP:conf/nips/YoonKDKBA18,DBLP:conf/iclr/GrantFLDG18,DBLP:conf/iclr/BaoWCB20}. (2) Metric-based methods, including Matching Network \cite{NIPS2016_90e13578}, PROTO \cite{DBLP:conf/nips/SnellSZ17}, Relation Network \cite{DBLP:conf/cvpr/SungYZXTH18}, TapNet \cite{DBLP:conf/icml/YoonSM19} and Induction Network \cite{DBLP:conf/emnlp/GengLLZJS19}, which aim to learn an appropriate distance metric to compare validation points with training points and make prediction through matching training points.

\paragraph{Transfer learning}~Few-shot text classification relates closely to transfer learning \cite{DBLP:journals/pieee/ZhuangQDXZZXH21} that aims to leverage knowledge from a related domain (a.k.a. source domain) to improve the learning performance and reduce the reliance on the number of labeled examples required in a target domain. Compared to meta-learning designed to aggregate the knowledge learned from many tasks, transfer learning typically involves a few tasks. In addition, we aim to directly reuse or fine-tune some existing representation in transfer learning, while a meta-learner is typically optimized at adapting to new tasks. Domain adaptation \cite{DBLP:journals/jmlr/GaninUAGLLML16,DBLP:conf/cvpr/TzengHSD17,DBLP:conf/iciot3/KhaddajH20} is a type of transfer learning, which aims to bridge the gap between the source and target domains by learning domain-invariant feature representations. Pre-trained model \cite{DBLP:conf/naacl/DevlinCLT19,DBLP:conf/nips/YangDYCSL19,DBLP:conf/nips/BrownMRSKDNSSAA20} can also be viewed as a type of transfer learning. The parameters pre-trained in the source domain are fine-tuned in the target domain, with faster training convergence.  
%which aims to share the parameters of the source learner to the target learner by fine-tuning techniques.

%Few-shot text classification is proposed to learn to recognize new classes with few training examples. 
\paragraph{Few-shot text classification}~To tackle few-shot text classification, a straightforward idea is to apply BERT \cite{DBLP:conf/naacl/DevlinCLT19} or XLNet \cite{DBLP:conf/nips/YangDYCSL19}, which have achieved strong performance in text classification by fine-tuning with a small number of training examples. Their performances can be less dependent on the number of training samples for the new classes.
%These pre-trained language models are less reliant  suggesting that pre-trained language models  with fine-tuning  may also be applied for few-shot text classification. 
Some other approaches are based on transfer learning. \citet{DBLP:journals/access/PanHGY19} proposed a modified hierarchical pooling strategy over pre-trained word embeddings to transfer knowledge obtained from some source domains to the target domain. \citet{DBLP:conf/coling/GuptaTO20} developed a binary classifier on the source domain to classify new classes by prefixing class identifiers to input texts.

Meta-learning \cite{DBLP:journals/corr/abs-1805-07722,DBLP:conf/naacl/YuGYCPCTWZ18,DBLP:conf/emnlp/GengLLZJS19,DBLP:conf/acl/GengLLSZ20,DBLP:conf/iclr/BaoWCB20} can also be utilized to solve few-shot text classification, and has achieved state-of-the-art performance. \citet{DBLP:conf/naacl/YuGYCPCTWZ18} proposed an adaptive metric learning approach that automatically determines the best weighted combination from meta-training tasks for few-shot tasks. \citet{DBLP:conf/emnlp/GengLLZJS19,DBLP:conf/acl/GengLLSZ20} leveraged the dynamic routing algorithm in meta-learning for few-shot text classification. \cite{DBLP:conf/iclr/BaoWCB20} leveraged distributional signatures (e.g. word frequency and information entropy) to train a model within a meta-learning framework. 

\begin{figure*}
    \centering
    \includegraphics[scale=0.45]{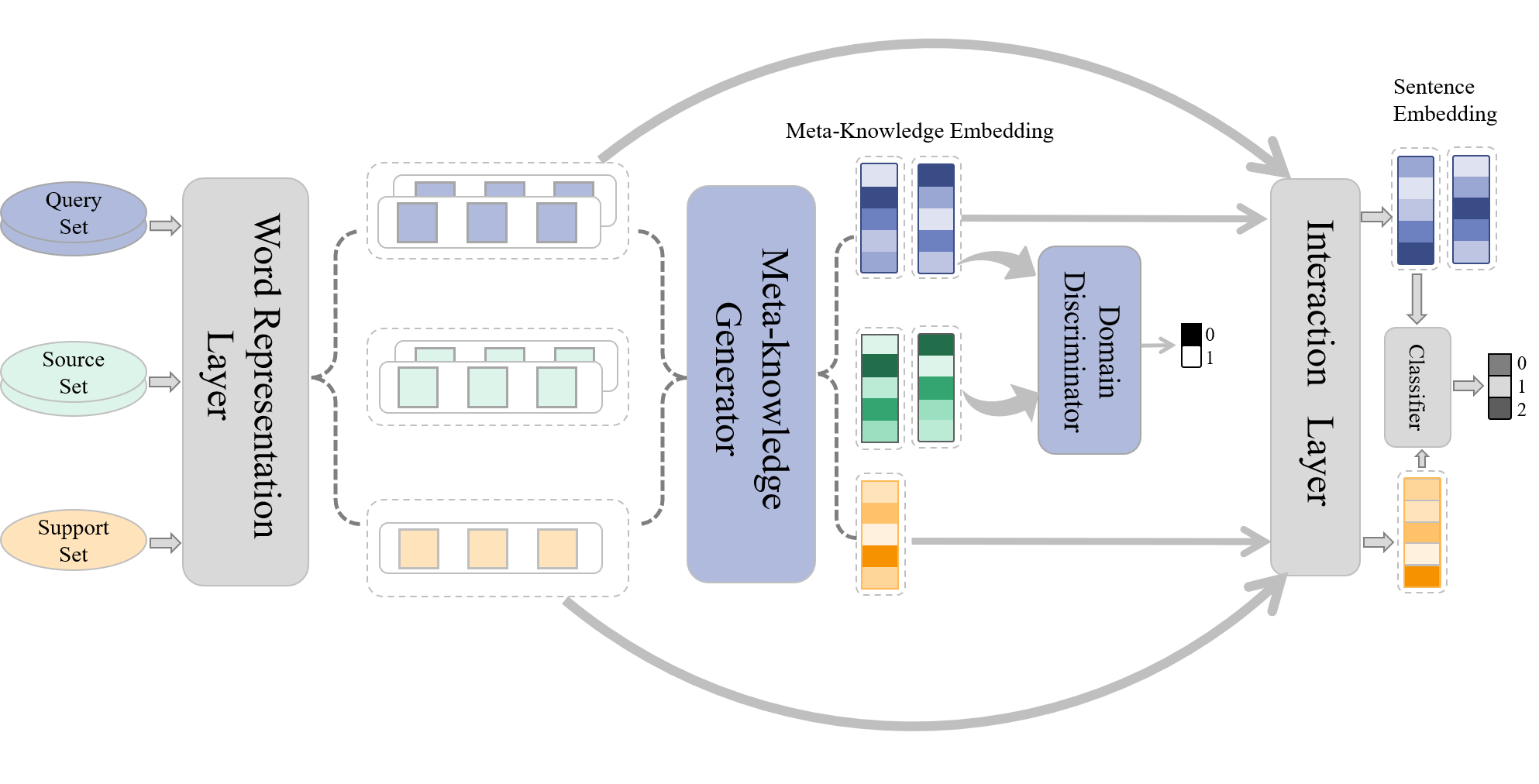}
    \caption{MLADA Network architecture for a $N$-way $K$-shot($N=3, K=2$) problem}
    \label{fig:model}
\end{figure*}

\section{Method}
\label{sec:Method}

% Since there are too few labeled examples in new classes to train a supervised classifier from scratch, performing meta-learning on the training set to extract task-agnostic knowledge is necessary. However, due to the complexity of natural language, only the general meta-learning architecture is insufficient to capture transferable features in natural language domain. Therefore, we aim to build a domain adversarial network, combined with the meta-learning training framework, for extracting more general transferable features.

In this section, we first present the preliminary background on episode-based meta-learning framework~ \cite{NIPS2016_90e13578}. After that, we explicitly describe the proposed MLADA (Meta-Learning Adversarial Domain Adaptation) Network.

\subsection{Episode-based meta-learning}

The goal of meta-training is to train a classifier that can learn meta-knowledge from training data. In this way, the classifier can quickly learn from a few annotations when classifying unseen classes. The ``episode'' training strategy that \citet{NIPS2016_90e13578} proposed has proved to be effective. The episode-based meta-learning consists of two main stages: 

\paragraph{Meta-training}

Firstly, \textit{N} classes are sampled from training data $\mathcal{Y}_{train}$. For each of these \textit{N} classes, two subsets of examples are sampled separately as the support set $S$ and the query set $Q$. Next, input the support set $S$ and the query set $Q$ to the model and update the parameters by minimizing the loss in the query set $Q$. The procedure above is called a training episode, which will be repeated multiple times.

\paragraph{Meta-testing}

After meta-training is finished, the performance of the model will be evaluated by the same episode-based mechanism. In a testing episode, $N$ new classes will be sampled from $\mathcal{Y}_{test}$, which is disjoint to $\mathcal{Y}_{train}$. Then the support set and the query set will be sampled from the $N$ classes. The model parameters can be fine-tuned through the small support set. The performance of the model will be evaluated through the average classification accuracy on the query set across all testing episodes.

We found that only a small subset of training data are accessible per training episode in the standard episode-based meta-training \cite{NIPS2016_90e13578}. To solve this problem, we build domain adversarial tasks to utilize more training data per training episode. Details of our model are described in the next section.

\subsection{Meta-Learning Adversarial Domain Adaptation Network (MLADA)}

\paragraph{Overview}

Our goal is to improve the performance of few-shot classification by combining adversarial domain adaptation and episode-based meta-learning. Figure \ref{fig:model} gives an overview
of our model. In the rest of this section, we will introduce the main components of the model.

\paragraph{Word Representation Layer}

The goal of this layer is to represent each word with a $d$-dimensional vector. Following \citet{DBLP:conf/iclr/BaoWCB20}, we construct the $d$-dimensional vector with the word embeddings, which is pre-trained with fastText \cite{DBLP:journals/corr/JoulinGBDJM16}.

\paragraph{Domain Discriminator}

We refer to the support set and the query set as the target domain and the rest of the training data as the source domain. We sample a subset of examples from the source domain as the source set. The goal of this module is to distinguish whether the sample is from the source domain or the target domain. The discriminator is a three layer feed-forward neural network. We apply the $softmax$ function in the output layer to evaluate the probability distribution $Pr(y|\lambda)$. $y = 0\ or\ 1$ represents that the sample is from the query set or the source set.

\paragraph{Meta-knowledge Generator}
\begin{algorithm}[htb]
 \caption{MLADA Training Procedure}
 \label{alg:MLADA}
 \begin{algorithmic}[1]
  \Require
  Training data $ \{\mathcal{X}_{train}, \mathcal{Y}_{train}\} $;
  $T$ episodes and $ep$ epochs;
  $N$ classes in support set or query set;
  $K$ samples in each class in the support set and $L$ samples in each class in the query set;
  The generator's parameters $\beta$, the discriminator's parameters $\mu$ and the classifier's parameters $\theta$.
  \Ensure
  Parameters $\beta$ and $\mu$ after training;
  \State Randomly initialize the model parameters $\beta$, $\mu$ and $\theta$;
  \label{code: Initialize}
  \For{each $i \in [1, ep]$}
  \State \vspace{-1.2em} \begin{enumerate}
      \item[] $\mathcal{Y} \leftarrow \Lambda(\mathcal{Y}_{train}, N)$;\footnotemark[1]
    %   \footnote{$\Lambda(\mathcal{Y}, N)$ denotes selecting $N$ elements from $\mathcal{Y}$ randomly.}
  \end{enumerate}
  \For {each $j \in [1, T]$}
    \State \vspace{-1.2em} \begin{enumerate}
        \item[] \begin{enumerate}
         \item[] $S, Q, \Phi \leftarrow \emptyset, \emptyset, \emptyset$;
        \end{enumerate}
    \end{enumerate}
    \For{$y \in \mathcal{Y}$}
    \State \vspace{-1.2em} \begin{enumerate}
      \item[] \begin{enumerate}
      \item[] \begin{enumerate}
        \item[] $S \leftarrow S \cup \Lambda(\mathcal{X}_{train}\{y\}, K)$;\footnotemark[2]
        % \footnote{$\mathcal{X}_{train}\{y\}$ denotes samples labeled $y$ in $\mathcal{X}_{train}$.}
        \end{enumerate}
        \end{enumerate}
    \end{enumerate}
    \State \vspace{-1.2em} \begin{enumerate}
      \item[] \begin{enumerate}
      \item[] \begin{enumerate}
        \item[] $Q \leftarrow Q \cup \Lambda(\mathcal{X}_{train}\{y\} \backslash S, L)$;
        \end{enumerate}
        \end{enumerate}
    \end{enumerate}
    \State \vspace{-1.2em} \begin{enumerate}
      \item[] \begin{enumerate}
      \item[] \begin{enumerate}
        \item[] $\Phi \leftarrow \Phi \cup \Lambda(\mathcal{X}_{train} \backslash \mathcal{X}_{train}\{y\}, L)$;
        \end{enumerate}
        \end{enumerate}
    \end{enumerate}
    \EndFor
    \State \vspace{-1.2em} \begin{enumerate}
      \item[] \begin{enumerate}
        \item[] Input $S$ to the model;
      \end{enumerate}
    \end{enumerate}
    \State \vspace{-1.2em} \begin{enumerate}
      \item[] \begin{enumerate}
        \item[] Fix $\mu,\beta$. Update $\theta$ by minimizing the Eq.\ref{loss_clf};
      \end{enumerate}
    \end{enumerate}
    \label{code:update clf parameters}
    \State \vspace{-1.2em} \begin{enumerate}
      \item[] \begin{enumerate}
        \item[] Input $Q, \Phi$ to the model;
      \end{enumerate}
    \end{enumerate}
    \State \vspace{-1.2em} \begin{enumerate}
      \item[] \begin{enumerate}
        \item[] Fix $\beta,\theta$. Update $\mu$ by minimizing the loss of the discriminator (Eq.\ref{loss_d});
      \end{enumerate}
    \end{enumerate}
    \label{code:update d parameters}
    \State \vspace{-1.2em} \begin{enumerate}
      \item[] \begin{enumerate}
        \item[] Fix $\mu, \theta$. Update $\beta$ by minimizing the loss of the generator (Eq.\ref{loss_g});
      \end{enumerate}
    \end{enumerate}
    % \State Fix the parameters $\mu$ and $\theta$. Input the source set $\Phi$ and the query set $Q$ to the model and update the parameters $\beta$ by minimizing the loss of the generator(as shown in the Eq.\ref{loss_g});
    \label{code:update g parameters}
  \EndFor
  \EndFor
% \RETURN $E_n$;
 \end{algorithmic}
\end{algorithm}

This module is mainly composed of a bi-directional LSTM (BiLSTM) and a fully connected layer. We utilize a BiLSTM to encode contextual embeddings for each time-step. The input of the module is a sequence of word vectors $P:[p_1, ...,p_m]$, where $m$ represents the number of words in a sentence. The output is a matrix $h^p_{d \times m}$, which is composed of contextual embeddings.
\begin{eqnarray}    \label{h_i^p}
\mathop{h_i^p}\limits^{\rightarrow} &=& \mathop{LSTM}\limits ^{\rightarrow}(\mathop{h_{i-1}^p}\limits ^{\rightarrow}, p_i) \quad i = 1, ...,m \\ 
\mathop{h_i^p}\limits^{\leftarrow} &=& \mathop{LSTM}\limits ^{\leftarrow}(\mathop{h_{i+1}^p}\limits ^{\leftarrow}, p_i) \quad i = m, ...,1 \\
h_i^p &=& Concat(\mathop{h_i^p}\limits^{\rightarrow}, \mathop{h_i^p}\limits^{\leftarrow}) \quad i = 1, ...,m  \\
h^p &=& [h_1^p, h_2^p, ..., h_m^p]
\end{eqnarray}
Next, we employ a single layer feed-forward neural network and apply the $softmax$ function to get the output $k^p$.
\begin{eqnarray}    \label{k^p}
k^p = Softmax(\omega \cdot h^p + b)
\end{eqnarray}
$k^p$ is an $n$-dimensional vector, which represents the meta-knowledge included in the sentence. $n$ denotes the length of the sentence.

The goal of the meta-knowledge generator is not only to make the final classification results better, but also to confuse the domain discriminator as much as possible, so that the discriminator can not distinguish between samples from query set or source set. The theory on domain adaptation suggests that, for effective domain transfer to be achieved, predictions must be made based on features that cannot discriminate between the source domain and target domain, which is the motivation for us to build the meta-knowledge generator.

\paragraph{Interaction Layer}

We consider that the vector generated by the meta-knowledge generator is the transferable features, and word embeddings is the specific features of sentences. The role of the interaction layer is to fuse transferable features and sentence-specific features to produce the output as sentence embeddings, which will be used as the input of the classifier to obtain the final classification results. Suppose that the length of the sentence $p$ is $m$, the word vectors is $w_i^p(i \in [1, m])$, the dimension of the word vector is $d$ and the meta-knowledge of the sentence is $k^p$, then the final sentence vector is $s^p$:
\begin{eqnarray}   \label{s_p}
s^p &=& W^p_{d \times m} \cdot k^p
\end{eqnarray}
where $W^p = [w_1^p, w_2^p, ..., w_m^p]$.

\paragraph{Classifier}

The classifier is trained by the support set from scratch for each episode. We choose the $ridge\ regression$ as the classifier. The reason why we adopt the ridge regression to fit the support set are as follows: 1) If we choose neural networks as the classifier, it will be trained inadequately because the number of samples in the support set is too small. 2) The ridge regression admits a closed-form solution and it reduces over-fitting on the small support set through proper regularization.Specifically, we minimize regularized squared loss:

\begin{small}
\begin{eqnarray}   \label{loss_clf}
% \hspace{-3mm}
\mathcal{L}^{RR}(\theta) = \frac{1}{2m}\sum_{i=1}^m[((f_\theta(x^{(i)}) - y^{(i)})^2 + \lambda\sum_{j=1}^n \theta_j^2)]
\end{eqnarray}
\end{small}
where $m$ represents the number of samples in the support set, $f_\theta(x^{(i)})$ represents the prediction of the ridge regressor, $y^{(i)}$ represents the label of the sample, $\sum_{j=1}^n \theta_j^2$ denotes the squared Frobenius norm and $\lambda>0$ controls the extent of the regularization.

\paragraph{Loss Function}

In each training episode, we first fix the parameters of the generator and the discriminator to update the classifier's parameters by the support set. The classifier's loss function is shown in Eq.\ref{loss_clf}.

Next, we fix the parameters of the generator and the classifier to update the discriminator's parameters by the query set and the source set. We use the cross-entropy loss as the discriminator's loss function, which is shown in Eq.\ref{loss_d}.
\begin{eqnarray}   \label{loss_d}
\mathcal{L}^{D}(\mu)\! &=&\! -\frac{1}{2m}\sum_{i=1}^{2m}[y_d^{(i)}log{D_\mu(k^{(i)})} \nonumber\\ 
\! &+&\! (1-y_d^{(i)})log(1-D_\mu(k^{(i)}))]
\end{eqnarray}
where $\mu$ denotes the parameters of the discriminator, $m$ represents the number of samples of the query set or the source set.$y_d = 0\ or\ 1$ denotes whether the sample is from the source set or the query set. $k$ represents the meta-knowledge vector.

\footnotetext[1]{$\Lambda(\mathcal{Y}, N)$ denotes selecting $N$ elements from $\mathcal{Y}$ randomly.}
\footnotetext[2]{$\mathcal{X}_{train}\{y\}$ denotes samples labeled $y$ in $\mathcal{X}_{train}$.}

Finally, we fix the parameters of the discriminator and the classifier to update the generator's parameters by the query set and the source set. The loss function of the generator is composed of two components. The first one is a cross-entropy loss for the final classification results, and the second one is the opposite of the discriminator's loss, which is to confuse the discriminator.
\begin{eqnarray}   \label{loss_g}
\mathcal{L}^{G}(\beta) = CELoss(f(W \cdot G_\beta(W)), y) - \mathcal{L}^D
\end{eqnarray}
where $\beta$ represents the generator's parameters. $f$ denotes the ridge regressor. $W$ represents the matrix of word vectors in a sentence. $y$ denotes the real labels of samples.$\mathcal{L}^D$ is shown in Eq.\ref{loss_d}.

\paragraph{Training Procedure}  It is remarkable that the meta-knowledge generator is optimized over all training episodes, while the classifier is trained from scratch for each episode. In each training episode, we first utilize the support set to update the parameters in the classifier. Next, we use the query set and source set to update the parameters of the meta-knowledge generator and the domain discriminator. The details of training procedure of our model are shown in Algorithm \ref{alg:MLADA}.

\section{Experiments}

In this section, we perform comprehensive experiments to compare our proposed model with five competitive baselines, and  evaluate the performance on four text classification datasets. %The experimental results fully demonstrate the effectiveness of our model.

\subsection{Datasets}
We use four benchmark datasets for text classification, whose statistics are summarized in Table~\ref{tab datasets}.
%We first provide an overview of the datasets in table \ref{tab datasets}. Next, we describe each dataset in detail.

\begin{table*}[htb]
\centering
\begin{tabular}{ccccc}
\hline
Dataset & Avg. text length & vocab size & \# samples & \# train / val / test classes\\
\hline
HuffPost & 11 & 8218 & 36900 & 20 / 5 / 16 \\
Amazon & 140 & 17062 & 24000 & 10 / 5 / 9 \\
Reuters & 168 & 2234 & 620 & 15 / 5 / 11 \\
20 Newsgroups & 340 & 32137 & 18820 & 8 / 5 / 7 \\
\hline
\end{tabular}
\caption{
Statistics of the four benchmark datasets. 
}
 \label{tab datasets}

\end{table*}

\renewcommand{\arraystretch}{1.1} %控制行高  
\begin{table*}[htb]  
  
  \centering  
%   \fontsize{10pt}\selectfont  
  \begin{threeparttable}  
    \begin{tabular}{ccccccccccc}  
    \toprule  
    \multirow{2}{*}{Method}&  
    \multicolumn{2}{c}{HuffPost}&\multicolumn{2}{c}{Amazon}&\multicolumn{2}{c}{Reuters}&\multicolumn{2}{c}{20 News}&\multicolumn{2}{c}{Average}\cr  
    \cmidrule(lr){2-3} \cmidrule(lr){4-5} \cmidrule(lr){6-7} \cmidrule(lr){8-9} \cmidrule(lr){10-11} 
    &1 shot&5 shot&1 shot&5 shot&1 shot&5 shot&1 shot&5 shot&1 shot&5 shot\cr  
    \midrule  
    MAML(2017) & 35.9 & 49.3 & 39.6 & 47.1 & 54.6 & 62.9 & 33.8 & 43.7 & 40.9 & 50.8 \\
    PROTO(2017) & 35.7 & 41.3 & 37.6 & 52.1 & 59.6 & 66.9 & 37.8 & 45.3 & 42.7 & 51.4 \\
    Induct(2019) & 38.7 & 49.1 & 34.9 & 41.3 & 59.4 & 67.9 & 28.7 & 33.3 & 40.4 & 47.9 \\
    HATT(2019) & 41.1 & 56.3 & 49.1 & 66.0 & 43.2 & 56.2 & 44.2 & 55.0 & 44.4 & 58.4 \\
    DS-FSL(2020) & 43.0 & 63.5 & 62.6 & 81.1 & 81.8 & 96.0 & 52.1 & 68.3 & 59.9 & 77.2 \\
    \midrule 
    MLADA(ours) & \bf45.0 & \bf64.9 & \bf68.4 & \bf86.0 & \bf82.3 & \bf96.7 & \bf59.6 & \bf77.8 & \bf63.9 & \bf81.4 \\ 
    \bottomrule  
    \end{tabular}  
    \caption{Mean accuracy (\%) of 5-way 1-shot and 5-way 5-shot classification over four datasets.}  
    \label{tab main experiment}  
    \end{threeparttable}  
\end{table*}

\noindent\textbf{HuffPost headlines} contains 41 classes of news headlines from the year 2012 to 2018 obtained from HuffPost \cite{dataset}. Its text is less abundant (i.e., with smaller text length) than the other datasets  and considered to be more challenging for text classification.
%and they are less grammatical than other datasets.

\noindent\textbf{Amazon product data} contains product reviews from $24$ product categories, including $142.8$ million reviews spanning 1996-2014 \cite{DBLP:conf/www/HeM16}.
Our task is to identify the product categories of the reviews. Since the original dataset is proverbially large, we sample a subset of $1,000$ reviews from each category.

\noindent\textbf{Reuters-21578} is collected from Reuters newswire in 1987. We use the standard ApteMode version of the dataset. Following \citet{DBLP:conf/iclr/BaoWCB20}, we consider 31 classes and remove multi-labeled articles. Each class contains at least 20 articles.

\noindent\textbf{20 Newsgroups} is a collection of approximately 20,000 newsgroup documents \cite{DBLP:conf/icml/Lang95}, partitioned (nearly) evenly across 20 different newsgroups. 

\subsection{Experiment Setup}

\paragraph{Baselines}
%We choose classical approaches in Few-shot learning and some latest methods for few-shot text classification as our baselines. The brief descriptions of these methods are as follow.
We compare our MLADA with multiple competitive baselines, which are briefly summarized in the following:
\begin{itemize}
\item \textbf{MAML}~\cite{DBLP:conf/icml/FinnAL17} is trained by maximizing the sensitivity of the loss functions of new tasks, so that it can rapidly adapt to new tasks after the parameters have been up-dated through few gradient steps.

\item \textbf{Prototypical Networks}~\cite{DBLP:conf/nips/SnellSZ17}, abbreviated as \textbf{PROTO},  is a metric-based method for few-shot classification by using sample averages as class prototypes.

\item \textbf{Induction Networks}~\cite{DBLP:conf/emnlp/GengLLZJS19} learns a class-wise representation by leveraging the dynamic routing algorithm in meta-learning.

\item \textbf{HATT}~\cite{DBLP:conf/aaai/GaoH0S19} extends \textbf{PROTO} by adding a hybrid attention mechanism to the prototypical network.

\item \textbf{DS-FSL}~\cite{DBLP:conf/iclr/BaoWCB20} is trained within a meta-learning framework to map the distribution signatures into attention scores so as to extract more transferable features.
\end{itemize}

\paragraph{Implementation Details}

\begin{figure*}[htb]
    \centering
    \subfigure[avg]{
    \label{fig:avg}
    \includegraphics[scale=0.32]{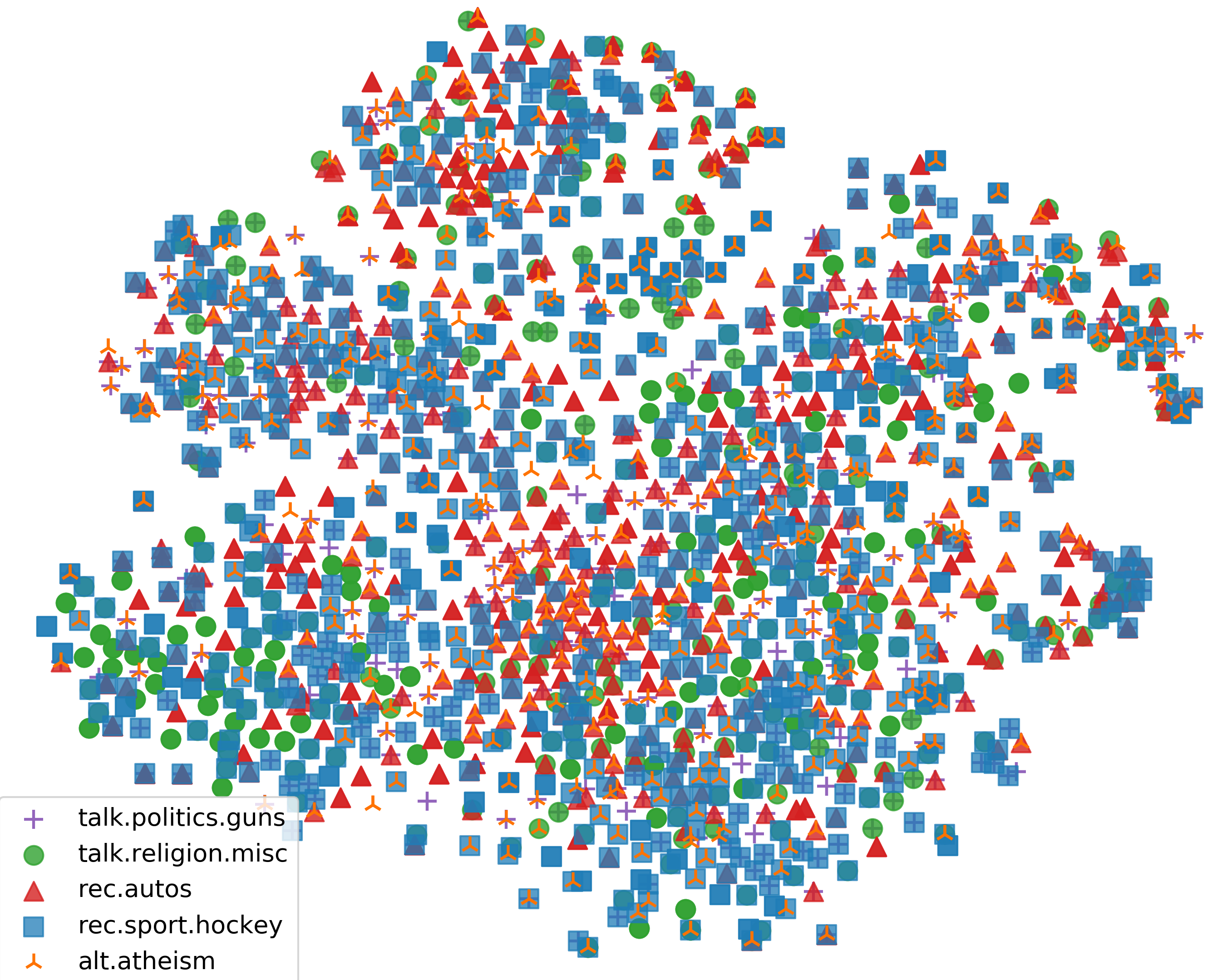}
    }
    \hspace{0.01in} % 两图片之间的距离
    \subfigure[DS-FSL(5-shot)]{
    \label{fig:ds}
    \includegraphics[scale=0.32]{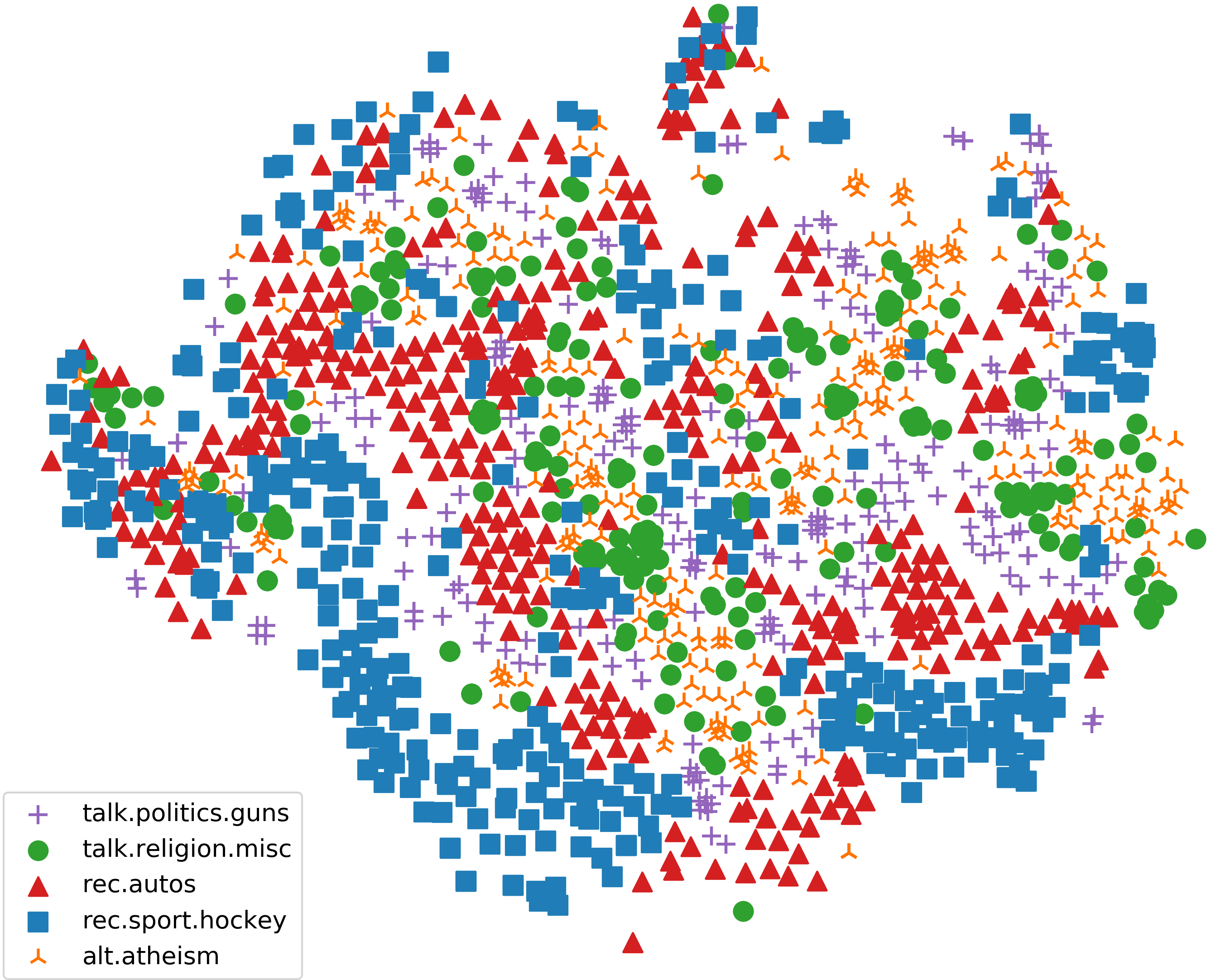}
    }
    \hspace{0.01in} % 两图片之间的距离
    \subfigure[MLADA(5-shot)]{
    \label{fig:MLADA}
    \includegraphics[scale=0.32]{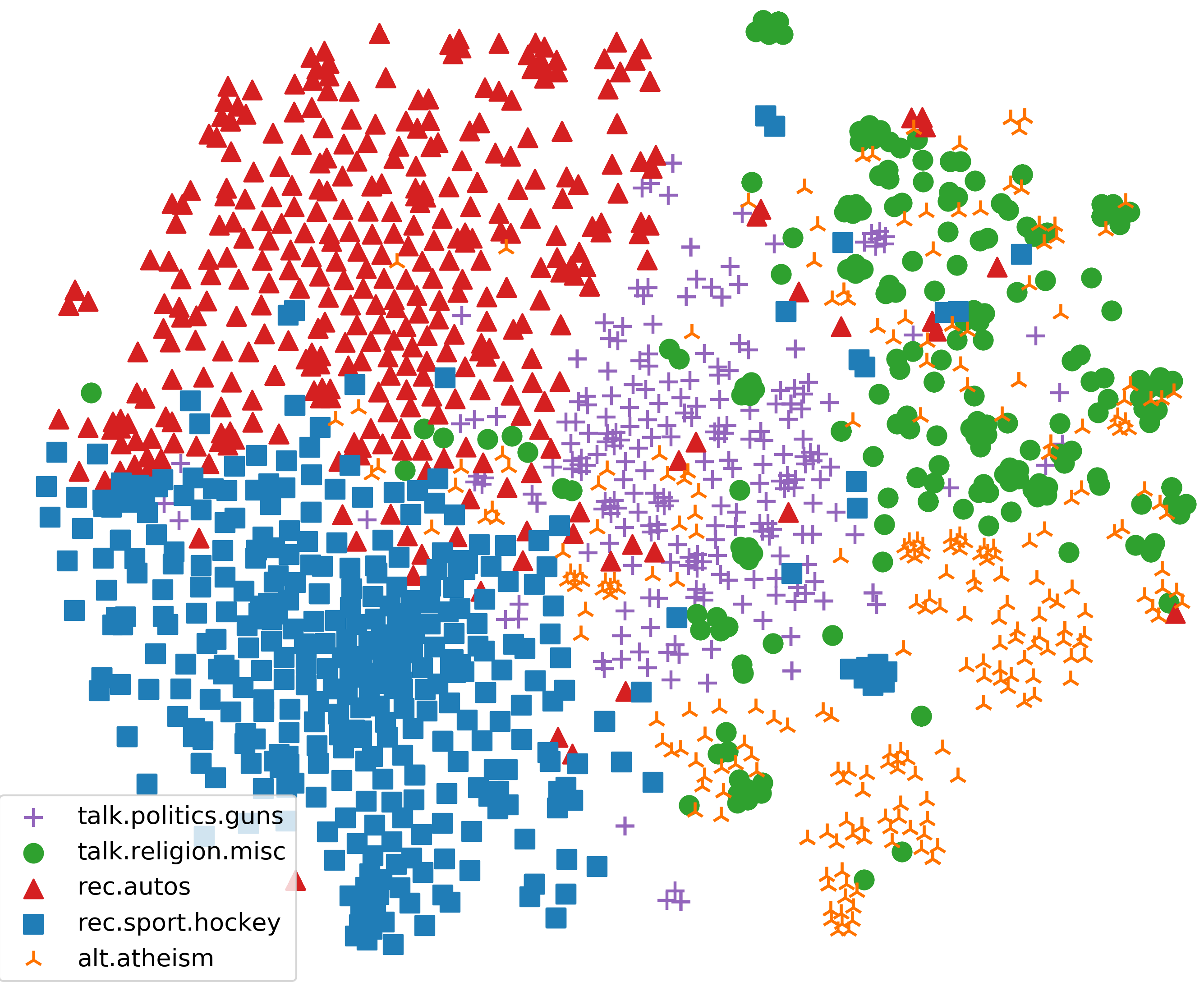}
    }\hspace{0.01in} % 两图片之间的距离
    \subfigure[MLADA(1-shot)]{
    \label{fig:MLADA_oneshot}
    \includegraphics[scale=0.32]{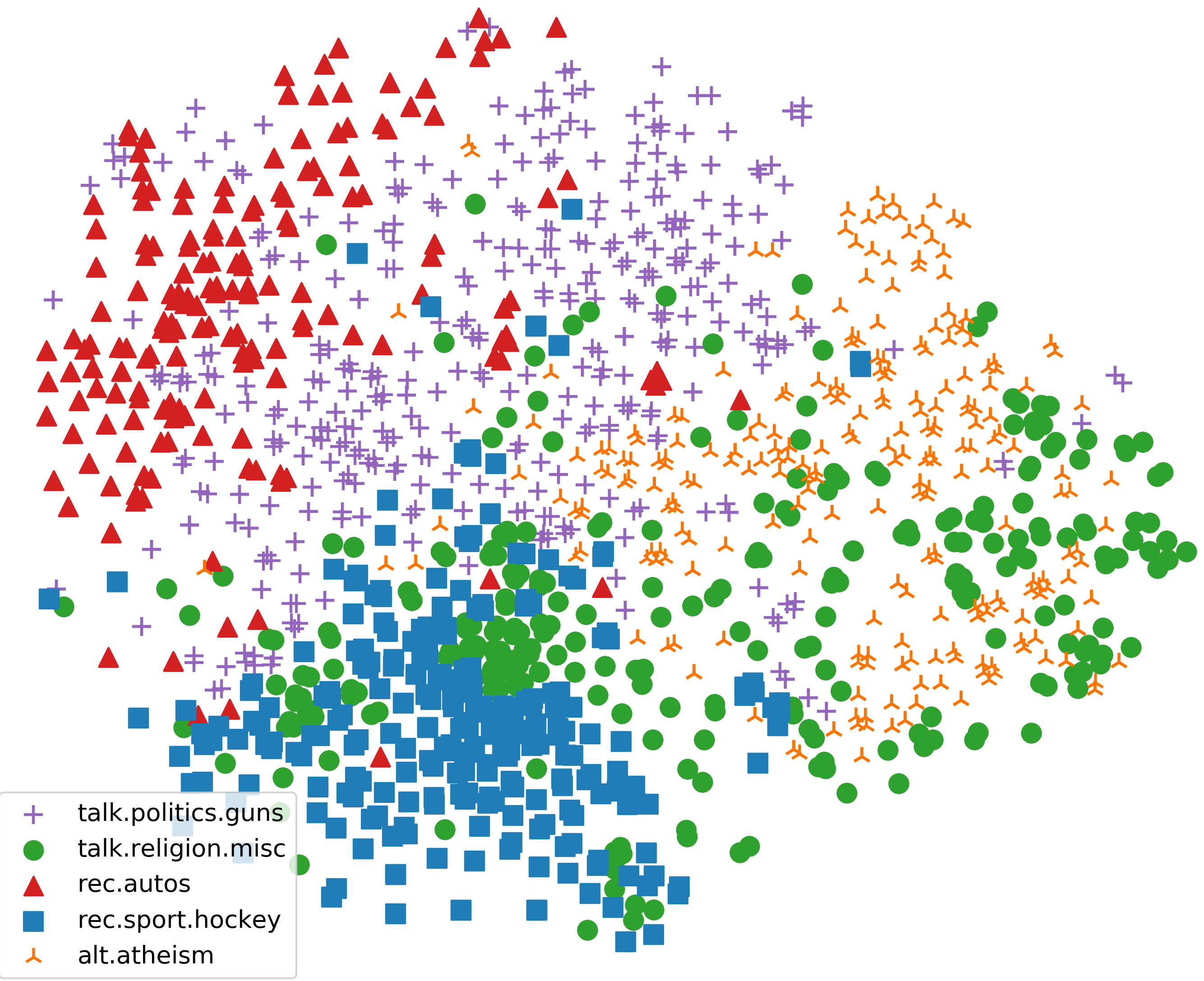}
    }
    \caption{t-SNE visualization of the input representation of the classifier for a testing episode($N=5$, $K=5$, $L=500$)sampled from 20 Newsgroups. Note that the 5 classes is not seen in training set. The input representation of the classifier given by (a) the average of word embeddings (b) DS-FSL and (c) MLADA(ours). (d) is the t-SNE visualization of MLADA on 5-way 1-shot classification.}
    \label{fig:analysis}
\end{figure*}

Following \citet{DBLP:conf/iclr/BaoWCB20}, we use pre-trained fastText~\cite{DBLP:journals/corr/JoulinGBDJM16} for word embedding. %For the out-of-vocabulary (OOV) words, we initialize the word embeddings randomly. 
In the meta-knowledge generator, we use a BiLSTM with $128$ hidden units.  In the domain discriminator, the numbers of hidden units for the two feed-forward layers are set to $256$ and $128$, respectively. All parameters are optimized using Adam with a learning rate of 0.001 \cite{DBLP:journals/corr/KingmaB14}.

During meta-training, we perform $100$ training episodes ($T=100$) per epoch. Meanwhile, we apply early stopping when the accuracy on the validation set fails to improve for $20$ epochs. We evaluate the model performance based on $1,000$ testing episodes and report the average accuracy over $5$ different random seeds. All the experiments are conducted on a NVIDIA v100 GPU.

\subsection{Experimental Results}

\begin{figure*}[htb]
\centering
\begin{tabular}{ccl}
\hline
Seen classes & \emph{Politics, Entertainment, Food\&Drink, College, Arts}  & Prediction \\
\hline
DS-FSL & \colorbox[RGB]{255,50,50}{Senate} \colorbox[RGB]{255,50,50}{committee} \colorbox[RGB]{255,200,200}{advances} \colorbox[RGB]{255,255,255}{bill} to \colorbox[RGB]{255,255,255}{protect} \colorbox[RGB]{255,200,200}{Robert} \colorbox[RGB]{255,200,200}{Mueller}. & \small{\emph{politics}} \cmark \\
MLADA(ours) & \colorbox[RGB]{255,50,50}{Senate} \colorbox[RGB]{255,100,100}{committee} \colorbox[RGB]{255,200,200}{advances} \colorbox[RGB]{255,200,200}{bill} to \colorbox[RGB]{255,200,200}{protect} \colorbox[RGB]{255,150,150}{Robert} \colorbox[RGB]{255,150,150}{Mueller}. & \small{\emph{politics}} \cmark \\
\hline
Unseen classes & \emph{Sports, Education, Media, Tech, Environment} & Prediction \\
\hline
DS-FSL & \colorbox[RGB]{255,200,200}{Olympic} \colorbox[RGB]{255,50,50}{committee} \colorbox[RGB]{255,200,200}{CEO} \colorbox[RGB]{255,150,150}{resigns} cites \colorbox[RGB]{255,100,100}{health} issues. & \small{\emph{environment}} \xmark \\
MLADA(ours) & \colorbox[RGB]{255,50,50}{Olympic} \colorbox[RGB]{255,200,200}{committee} \colorbox[RGB]{255,200,200}{CEO} \colorbox[RGB]{255,200,200}{resigns} cites \colorbox[RGB]{255,200,200}{health} issues. & \small{\emph{sports}} \cmark \\
\hline
\end{tabular}
\caption{ \label{fig:attention}
The visualization of attention weights generated by DS-FSL and the meta-knowledge generator of our model.
}
\end{figure*}

The experimental results are reported in Table \ref{tab main experiment}. Our model achieves the best performance across all datasets, with an average accuracy of $63.9\%$  in 1-shot classification and $81.4\%$ in 5-shot classification, outperforming the state-of-the-art model DS-FSL \cite{DBLP:conf/iclr/BaoWCB20} by a notable $4\%$ improvement. For DS-FSL, it extracts transferable features via certain distribution signatures (e.g., word frequency or information entropy), but ignores other information of sentences, including implicit
interaction between words. In contrast, we does not limit the transferable knowledge to statistical information. Our strategy is to combine the proposed domain adversarial network with meta-learning, generating more comprehensive transferable features.

Furthermore, our model improves dramatically $7.5\%$ and $9.5\%$ on 20 Newsgroups in 1-shot and 5-shot classification. The average length of texts in the 20 Newsgroups is longer than the other datasets. The empirical results clearly demonstrate that our model is more suitable for longer texts, which contain more abundant text information.

\subsection{Ablation Study}

We conduct an ablation study to examine the effectiveness of the proposed domain adversarial network as well as the interaction layer and the source set. The results of Amazon dataset are reported in Table \ref{tab ablation}.

Firstly, we use a bi-directional LSTM instead of the proposed domain adversarial network (including the meta-knowledge generator and the domain discriminator) for sentence encoding. The performances in the tasks of 1-shot classification and 5-shot classification decrease by 6.5\% and 5.3\%, respectively. This verifies the effectiveness of the proposed domain adversarial network. 

Secondly, we study how the interaction layer contributes to the performance of our model. We concatenate the vector generated by the meta-knowledge generator directly with the average sentence embedding instead of the interaction layer. From the result in Table~\ref{tab ablation}, we can see that our proposed interaction layer to combine the transferable features with the sentence-specific information are indeed more effective. 

Finally, we remove the source set and utilize the discriminator to distinguish the true classes of samples. We observe that the source set is also important to performance. Due to the removal of the source set, the model has only access to the support set and the query set in each training episode. Therefore, it cannot learn cross-domain transferable features.

\renewcommand{\arraystretch}{1.1} %控制行高  
\begin{table}[htb]  
  
  \centering  
%   \fontsize{10pt}\selectfont  
  \begin{threeparttable}  
    \begin{tabular}{lcc}  
    \toprule  
    \multirow{2}{*}{Models}&  
    \multicolumn{2}{c}{Accuracy(\%)}\cr  
    \cmidrule(lr){2-3} 
    &1 shot&5 shot\cr  
    \midrule  
    $-$ Domain Adversarial Network & 61.9 & 80.7 \\
    $-$ Interaction Layer & 66.6 & 83.0 \\
    $-$ Source Set & 67.1 & 84.2 \\
    \midrule 
    MLADA & \bf68.4 & \bf86.0 \\ 
    \bottomrule  
    \end{tabular}  
    \caption{Ablation study results of 5-way 1-shot and 5-way 5-shot classification on the Amazon dataset.}  
    \label{tab ablation}  
    \end{threeparttable}  
\end{table} 

\subsection{Visualization}

We utilize visualization experiments to demonstrate that our model can generate high-quality sentence embeddings and identify important lexical features for unseen classes.

We first use t-SNE \cite{van2008visualizing} visualization of sentence embeddings generated by different methods on the query set, as shown in Figure\ref{fig:analysis}. Compared to \ref{fig:avg} average word embeddings and \ref{fig:ds} DS-FSL, our method produces better separation both in 1-shot and 5-shot classification, demonstrating the effectiveness of MLADA in leveraging the supervised learning experience to generate high-quality sentence embeddings for few-shot text classification.

Moreover, we visualize the weight vectors generated by the meta-knowledge generator and compare it with DS-FSL, as shown in Figure \ref{fig:attention}. Our model reduces the weight of “committee” while increasing the weight of “Olympic”, which demonstrates that our model can recognize important lexical features in the new task, rather than simply transferring features obtained from experience.

\section{Conclusion}

In this paper, we propose a novel meta-learning approach called Meta-Learning Adversarial Domain Adaptation Network(MLADA), which can recognize important lexical features and generate high-quality sentence embeddings in new classes(not seen in training data). Specifically, we design an adversarial domain adaptation network in meta-training episodes, which aims to extract domain-invariant features and improve the adaptability of the meta-learner in new classes. We demonstrate that our method outperforms the existing state-of-the-art approaches on four standard text classification datasets. Future work includes applying MLADA to other fields including computer vision and speech recognition, and exploring the combination between adversarial domain adaptation network and other FSL algorithms.
 
\section{Acknowledgments}
\label{sec:acknowledge}

This work has been supported by the National Key Research and Development Program of China under Grant 2016YFB1000905, the National Natural Science Foundation of China under Grant No. U1811264, U1911203, 61877018, 61672234, 61672384 and Alibaba Group through Alibaba Innovative Research Program.

\bibliographystyle{acl_natbib}
\bibliography{anthology,acl2021}

\end{document}